\PassOptionsToPackage{table}{xcolor}
\documentclass[sigconf=true, nonacm=true]{acmart}

\usepackage{booktabs} % For formal tables
\usepackage{colortbl} % colored hline

\usepackage{etoolbox}
\usepackage{amsmath}
\usepackage{amsthm}
\usepackage{xspace}
\usepackage[table]{xcolor}
\usepackage{dsfont}
\usepackage{rotating}
\usepackage{bm}

\usepackage[algo2e,ruled,vlined,linesnumbered]{algorithm2e}
\usepackage{url}

\copyrightyear{2021}
\acmYear{2021}
\setcopyright{none}
\acmConference[GECCO '21]{Genetic and Evolutionary Computation Conference}{July 10--14, 2021}{Lille, France}
\acmBooktitle{Genetic and Evolutionary Computation Conference (GECCO '21), July 10--14, 2021, Lille, France}
\acmPrice{15.00}
\acmDOI{10.1145/nnnnnnn.nnnnnnn}
\acmISBN{978-x-xxxx-xxxx-x/YY/MM}

\DeclareMathOperator{\Bin}{Bin}
\newcommand{\R}{\mathbb{R}}

\newcommand{\opl}{$(1 + \lambda)$~EA\xspace}

\newcommand{\om}{\textsc{OneMax}\xspace}
\newcommand{\OM}{\textsc{OM}\xspace}
\newcommand{\onemax}{\textsc{OneMax}\xspace}

\newcommand{\oplheader}{\texorpdfstring{$(1+\lambda)$}{(1+lambda)}}

\DeclareMathOperator{\select}{select}
\DeclareMathOperator{\SBM}{SBM}
\DeclareMathOperator{\flip}{flip}
\newcommand{\shift}{0 \rightarrow 1}
\newcommand{\flipOp}[1]{\flip_{#1}}

\allowdisplaybreaks

\usepackage{tikz}
\usetikzlibrary{positioning,external}
\usepackage{pgfplots}
\pgfplotsset{compat=newest}

\usepackage{pgfplotstable}

\usepackage{mathtools}

\pgfplotsset{select coords between index/.style 2 args={
    x filter/.code={
        \ifnum\coordindex<#1\fi
        \ifnum\coordindex>#2\fi
    }
}}

\pgfplotscreateplotcyclelist{genericcycle}{%
mark options={scale=0.5},mark=*,blue\\
mark options={scale=0.5},mark=*,violet!80!black\\
mark options={scale=0.5},mark=*,red!80!black\\
mark options={scale=0.5},mark=*,orange\\
mark options={scale=0.5},mark=*,yellow!50!black\\
mark options={scale=0.5},mark=*,green!70!black\\
mark options={scale=0.5},mark=*,cyan!80!black\\
mark options={scale=0.5},mark=*,magenta!80!black\\
mark options={scale=0.5},mark=*,black\\
}

\pgfplotscreateplotcyclelist{runtimecycle}{%
mark options={scale=0.5},mark=*,blue\\             mark options={scale=0.5},mark=*,red!80!black\\             mark options={scale=0.5},mark=*,black\\
mark options={scale=0.5},mark=*,violet!80!black\\  mark options={scale=0.5},mark=*,orange\\                   mark options={scale=0.5},mark=*,gray\\
mark options={scale=0.5},mark=*,cyan!80!black\\    mark options={scale=0.5},mark=*,yellow!50!black\\          mark options={scale=0.5},mark=*,brown\\
mark options={scale=0.5},mark=*,magenta!80!black\\
}

\pgfplotscreateplotcyclelist{overlaidcycle}{%
blue\\
thick,dashed,blue\\
violet!80!black\\
thick,dashed,violet!80!black\\
red!80!black\\
thick,dashed,red!80!black\\
orange\\
thick,dashed,orange\\
yellow!50!black\\
thick,dashed,yellow!50!black\\
green!70!black\\
thick,dashed,green!70!black\\
cyan!80!black\\
thick,dashed,cyan!80!black\\
magenta!80!black\\
thick,dashed,magenta!80!black\\
black\\
thick,dashed,black\\
}

\pgfplotscreateplotcyclelist{withbincycle}{%
mark options={scale=0.5},mark=*,blue\\
mark options={scale=0.5},mark=*,violet!80!black\\
mark options={scale=0.5},mark=*,red!80!black\\
mark options={scale=0.5},mark=*,orange\\
mark options={scale=0.5},mark=*,green!70!black\\
very thick,dashed,magenta!80!black\\
}

\pgfplotscreateplotcyclelist{upperlowercycle}{%
mark options={scale=0.5},mark=*,violet!80!black\\
mark options={scale=0.5},mark=*,red!80!black\\
mark options={scale=0.5},mark=*,orange\\
very thick,dashed,blue\\
very thick,dashed,green!70!black\\
}

\newcommand{\FitnessLogSixteen}{static-results-v2/static-16-fitness-log-short.csv}
\pgfplotstableread[col sep=comma]{static-results-v2/static-3-deviations.csv}\DeviationsThree
\pgfplotstableread[col sep=comma]{static-results-v2/static-100-deviations.csv}\DeviationsHundred
\pgfplotstableread[col sep=comma]{static-results-v2/static-wallclock.csv}\StaticWallClock
\pgfplotstableread[col sep=comma]{static-results-v2/static-16-distributions-0.csv}\DistributionsSixteenZero
\pgfplotstableread[col sep=comma]{static-results-v2/static-11-distributions-0.csv}\DistributionsElevenZero
\pgfplotstableread[col sep=comma]{static-results-v2/static-k1-distributions-0.csv}\OneBitFlipsTable
\pgfplotstableread[col sep=comma]{static-results-v2/static-k2-distributions-0.csv}\TwoBitFlipsTable
\pgfplotstableread[col sep=comma]{static-results-v2/algorithm-expectations-flat.csv}\RuntimeComparisonTable

\newcommand{\ConvergenceCurves}{\begin{tikzpicture}
    \begin{semilogyaxis}[width=\linewidth, height=35ex, grid=both, minor x tick num=5, minor y tick num=4, legend cell align={left}, cycle list name=genericcycle,
                         filter discard warning=false, unbounded coords=jump, xlabel={Fitness evaluations}, ylabel={$\text{Fitness} - 10.022522$}]
        \addplot+ table[x=evaluation,y expr={\thisrow{lambda}==8 && \thisrow{attempt}==0 ? \thisrow{fitness}-10.022522 : NaN},col sep=comma] {\FitnessLogSixteen};
        \addplot+ table[x=evaluation,y expr={\thisrow{lambda}==8 && \thisrow{attempt}==1 ? \thisrow{fitness}-10.022522 : NaN},col sep=comma] {\FitnessLogSixteen};
        \addplot+ table[x=evaluation,y expr={\thisrow{lambda}==8 && \thisrow{attempt}==2 ? \thisrow{fitness}-10.022522 : NaN},col sep=comma] {\FitnessLogSixteen};
        \addplot+ table[x=evaluation,y expr={\thisrow{lambda}==8 && \thisrow{attempt}==3 ? \thisrow{fitness}-10.022522 : NaN},col sep=comma] {\FitnessLogSixteen};
    \end{semilogyaxis}
\end{tikzpicture}}

\newcommand{\ConvergencePrecisionThree}{\begin{tikzpicture}
    \begin{semilogyaxis}[width=0.59\linewidth, xmin=1, ymin=1e-17, ymax=30, height=35ex, enlarge x limits=0.04, cycle list name=genericcycle,
                         grid=both, minor x tick num=5, minor y tick num=3, legend pos=north west, legend cell align={left},
                         filter discard warning=false, unbounded coords=jump,
                         xlabel={Run number, $n=3$}, ylabel={Fitness difference to best}]
        \addplot+ table[x expr={\thisrow{lambda}==1 ? \thisrow{index}+1 : NaN},y=relative deviation] {\DeviationsThree};
        \addlegendentry{\small $\lambda=1$};
        \addplot+ table[x expr={\thisrow{lambda}==8 ? \thisrow{index}+1 : NaN},y=relative deviation] {\DeviationsThree};
        \addlegendentry{\small $\lambda=8$};
        \addplot+ table[x expr={\thisrow{lambda}==32 ? \thisrow{index}+1 : NaN},y=relative deviation] {\DeviationsThree};
        \addlegendentry{\small $\lambda=32$};
        \addplot+ table[x expr={\thisrow{lambda}==256 ? \thisrow{index}+1 : NaN},y=relative deviation] {\DeviationsThree};
        \addlegendentry{\small $\lambda=256$};
    \end{semilogyaxis}
\end{tikzpicture}}

\newcommand{\ConvergencePrecisionHundred}{\begin{tikzpicture}
    \begin{semilogyaxis}[width=0.59\linewidth, xmin=1, ymin=1e-17, ymax=30, yticklabels=\empty, height=35ex, enlarge x limits=0.04, cycle list name=genericcycle,
                         grid=both, minor x tick num=5, minor y tick num=3, legend pos=north west, legend cell align={left},
                         filter discard warning=false, unbounded coords=jump,
                         xlabel={Run number, $n=100$}]
        \addplot+ table[x expr={\thisrow{lambda}==6 ? \thisrow{index}+1 : NaN},y=relative deviation] {\DeviationsHundred};
        \addlegendentry{\small $\lambda=6$};
        \addplot+ table[x expr={\thisrow{lambda}==8 ? \thisrow{index}+1 : NaN},y=relative deviation] {\DeviationsHundred};
        \addlegendentry{\small $\lambda=8$};
        \addplot+ table[x expr={\thisrow{lambda}==32 ? \thisrow{index}+1 : NaN},y=relative deviation] {\DeviationsHundred};
        \addlegendentry{\small $\lambda=32$};
        \addplot+ table[x expr={\thisrow{lambda}==256 ? \thisrow{index}+1 : NaN},y=relative deviation] {\DeviationsHundred};
        \addlegendentry{\small $\lambda=256$};
    \end{semilogyaxis}
\end{tikzpicture}}

\newcommand{\ConvergenceInDistributions}{\begin{tikzpicture}
    \begin{loglogaxis}[width=\linewidth, height=34ex, grid=both, legend pos=south east, legend columns=2, legend cell align={left}, cycle list name=genericcycle,
                         filter discard warning=false, unbounded coords=jump, xlabel={$\lambda$}, ylabel={Max stddev}]
        \addplot+ table[x=lambda, y=maxDeviation,col sep=comma] {static-results-v2/static-3-distribution-stddev.csv};
        \addlegendentry{\small $n=3$};
        \addplot+ table[x=lambda, y=maxDeviation,col sep=comma] {static-results-v2/static-11-distribution-stddev.csv};
        \addlegendentry{\small $n=11$};
        \addplot+ table[x=lambda, y=maxDeviation,col sep=comma] {static-results-v2/static-5-distribution-stddev.csv};
        \addlegendentry{\small $n=5$};
        \addplot+ table[x=lambda, y=maxDeviation,col sep=comma] {static-results-v2/static-16-distribution-stddev.csv};
        \addlegendentry{\small $n=16$};
        \addplot+ table[x=lambda, y=maxDeviation,col sep=comma] {static-results-v2/static-8-distribution-stddev.csv};
        \addlegendentry{\small $n=8$};
        \addplot+ table[x=lambda, y=maxDeviation,col sep=comma] {static-results-v2/static-100-distribution-stddev.csv};
        \addlegendentry{\small $n=100$};
    \end{loglogaxis}
\end{tikzpicture}}

\newcommand{\WallClockTimes}{\begin{tikzpicture}
    \begin{loglogaxis}[width=\linewidth, height=35ex, grid=both, legend pos=north west, legend cell align={left}, cycle list name=upperlowercycle,
                         filter discard warning=false, xlabel={$n$}, ylabel={Wall-clock time},
                         domain=32:100, samples=10]
        \addplot+ table[x=n,y expr={\thisrow{lambda}==1 ? \thisrow{wall-clock time} : NaN}] {\StaticWallClock};
        \addlegendentry{\small $\lambda=1$}
        \addplot+ table[x=n,y expr={\thisrow{lambda}==32 ? \thisrow{wall-clock time} : NaN}] {\StaticWallClock};
        \addlegendentry{\small $\lambda=32$}
        \addplot+ table[x=n,y expr={\thisrow{lambda}==1024 ? \thisrow{wall-clock time} : NaN}] {\StaticWallClock};
        \addlegendentry{\small $\lambda=1024$}
        \addplot+ {2e-6 * x^4};
        \addlegendentry{\small $2 \cdot 10^{-6} \cdot n^4$};
        \addplot+ {4e-6 * x^4.2};
        \addlegendentry{\small $4 \cdot 10^{-6} \cdot n^{4.2}$};
    \end{loglogaxis}
\end{tikzpicture}}

\newcommand{\DistributionsByLambdasSixteenX}[1]{
    \addplot+ table[x=lambda,y expr={\thisrow{distance}==#1 ? \thisrow{probability} : NaN}] {\DistributionsSixteenZero};
    \addlegendentry{\small $k=#1$};
}
\newcommand{\DistributionsByLambdasSixteen}{\begin{tikzpicture}
    \begin{semilogxaxis}[width=\linewidth, height=45ex, grid=both, legend pos=north east, legend columns=2, legend cell align={left}, legend style={name=thelegend},
                         filter discard warning=false, xlabel={$\lambda$}, ylabel={Probability}, cycle list name=genericcycle]
        \DistributionsByLambdasSixteenX{1}           \DistributionsByLambdasSixteenX{10}
        \DistributionsByLambdasSixteenX{2}           \DistributionsByLambdasSixteenX{12}
        \DistributionsByLambdasSixteenX{3}           \DistributionsByLambdasSixteenX{14}
        \DistributionsByLambdasSixteenX{8}           \DistributionsByLambdasSixteenX{16}
        \DistributionsByLambdasSixteenX{9}
    \end{semilogxaxis}
\end{tikzpicture}}

\newcommand{\DistributionsByDistanceElevenX}[1]{
    \addplot+ table[x=distance,y expr={\thisrow{lambda}==#1 ? \thisrow{probability} : NaN}] {\DistributionsElevenZero};
    \addlegendentry{\small $\lambda=#1$};
}
\newcommand{\DistributionsByDistanceEleven}{\begin{tikzpicture}
    \begin{axis}[width=\linewidth, height=45ex, grid=both, legend pos=north east, legend cell align={left}, legend style={name=thelegend}, cycle list name=withbincycle,
                 filter discard warning=false, xlabel={$k$}, ylabel={Probability}]
        \DistributionsByDistanceElevenX{64}
        \DistributionsByDistanceElevenX{128}
        \DistributionsByDistanceElevenX{256}
        \DistributionsByDistanceElevenX{512}
        \DistributionsByDistanceElevenX{1024}
        \addplot+ coordinates {
            (1,  0.005373717635564241)
            (2,  0.026868588177821204)
            (3,  0.08060576453346362)
            (4,  0.16121152906692723)
            (5,  0.2256961406936981)
            (6,  0.22569614069369812)
            (7,  0.16121152906692723)
            (8,  0.08060576453346362)
            (9,  0.026868588177821204)
            (10, 0.005373717635564241)
            (11, 4.885197850512946E-4)
        };
        \addlegendentry{\small $\Bin_{>0}(11, \frac{1}{2})$};
    \end{axis}
\end{tikzpicture}}

\newcommand{\DistributionsThreeNoise}{\begin{tikzpicture}
\begin{axis}[xbar stacked, bar width=0.8pt, width=0.95\linewidth, height=25ex, legend pos=outer north east, xmin=0, xmax=1, enlarge x limits=false, enlarge y limits=false]
\addplot[blue!60!white,fill=blue!60!white] coordinates {(0.06521466313931668,1)(0.15326420861865658,2)(0.15700454898041177,3)(0.15749086153904468,4)(0.16412168434798888,5)(0.18401993855492005,6)(0.19591574157125718,7)(0.2077802769365284,8)(0.2222580495125056,9)(0.2229283756621394,10)(0.2630507368648733,11)(0.272336538294655,12)(0.2801998482051622,13)(0.2863811908460284,14)(0.28851043355517225,15)(0.2947537131564156,16)(0.30247925176248674,17)(0.30498042098601014,18)(0.3056653160444334,19)(0.31101176492819044,20)(0.321466156748477,21)(0.32517980807475755,22)(0.32813343059040545,23)(0.3511916259667216,24)(0.3534118976226117,25)(0.3686570284855785,26)(0.3721691400415881,27)(0.3827742859496349,28)(0.38602750289556126,29)(0.40540884244217434,30)(0.41979637195865355,31)(0.42712233300058566,32)(0.4331176478362107,33)(0.4451123988760386,34)(0.4512314410805968,35)(0.4548690577048833,36)(0.45917266685363806,37)(0.46039718472636615,38)(0.46882837162179186,39)(0.4746668008032729,40)(0.4882270784881313,41)(0.49542329834088616,42)(0.49964712816326234,43)(0.5086329089621218,44)(0.521157900582887,45)(0.5301584505089488,46)(0.5416773224636621,47)(0.5453794302803254,48)(0.5559203253305924,49)(0.7982261468742895,50)};
\addlegendentry{\small $k=1$};
\addplot[red!60!white,fill=red!60!white] coordinates {(0.5787112523040906,1)(0.3205761253842076,2)(0.7101402730444272,3)(0.6547583036367618,4)(0.2471654345059855,5)(0.29320752808727335,6)(0.3287364994189298,7)(0.13723943193592691,8)(0.40613674537843486,9)(0.38350930584304843,10)(0.6804959584790067,11)(0.4969162525960419,12)(0.24365915474747063,13)(0.11659446477760782,14)(0.16650587377665516,15)(0.11716733169043018,16)(0.5822937695308503,17)(0.23016017329816812,18)(0.5391925224371331,19)(0.544054893383764,20)(0.3068674980804213,21)(0.06726003687202889,22)(0.28369210913234844,23)(0.30545586016586673,24)(0.2508519557150251,25)(0.5695641577573196,26)(0.4222613940659276,27)(0.3320902640076491,28)(0.4482314417140205,29)(0.355575849328426,30)(0.47416532319413146,31)(0.38036511014852964,32)(0.2972359720254017,33)(0.518624267772469,34)(0.3751877696525821,35)(0.14807961972354497,36)(0.3004440762813184,37)(0.3355607361713361,38)(0.2789813263843789,39)(0.1909519242068119,40)(0.31719390422131305,41)(0.4382122684881717,42)(0.30361008023787095,43)(0.337825405450127,44)(0.3815567633546341,45)(0.3913526556820629,46)(0.11051572829062069,47)(0.18577366895770195,48)(0.110632714075186,49)(0.11129626457731985,50)};
\addlegendentry{\small $k=2$};
\addplot[violet!60!white, fill=violet!60!white] coordinates {(0.3560740845565928,1)(0.5261596659971358,2)(0.13285517797516114,3)(0.18775083482419366,4)(0.5887128811460257,5)(0.5227725333578066,6)(0.47534775900981313,7)(0.6549802911275447,8)(0.37160520510905937,9)(0.3935623184948121,10)(0.05645330465611987,11)(0.23074720910930313,12)(0.4761409970473673,13)(0.5970243443763638,14)(0.5449836926681727,15)(0.5880789551531542,16)(0.11522697870666303,17)(0.46485940571582174,18)(0.15514216151843357,19)(0.14493334168804564,20)(0.3716663451711018,21)(0.6075601550532136,22)(0.38817446027724617,23)(0.3433525138674117,24)(0.39573614666236323,25)(0.06177881375710183,26)(0.2055694658924844,27)(0.2851354500427161,28)(0.1657410553904183,29)(0.23901530822939968,30)(0.10603830484721509,31)(0.1925125568508847,32)(0.26964638013838765,33)(0.03626333335149261,34)(0.17358078926682102,35)(0.3970513225715719,36)(0.2403832568650434,37)(0.20404207910229777,38)(0.25219030199382936,39)(0.33438127498991527,40)(0.19457901729055566,41)(0.06636443317094223,42)(0.19674279159886673,43)(0.15354168558775133,44)(0.0972853360624787,45)(0.07848889380898841,46)(0.34780694924571737,47)(0.26884690076197276,48)(0.3334469605942217,49)(0.09047758854839073,50)};
\addlegendentry{\small $k=3$};
\end{axis}
\end{tikzpicture}}

\newcommand{\OneAndTwoBitFlipsX}[3]{
    \addplot+ table[x=lambda,y expr={\thisrow{n}==#1 ? \thisrow{probability} : NaN}] {#3};
    \addlegendentry{\scriptsize $n=#1, k=#2$};
}

\newcommand{\OneAndTwoBitFlips}{\begin{tikzpicture}
    \begin{semilogxaxis}[width=0.77\linewidth, height=42ex, grid=both, legend pos=outer north east, legend cell align={left}, legend style={name=thelegend}, cycle list name=overlaidcycle,
                 filter discard warning=false, xlabel={$\lambda$}, ylabel={Probability}]
        \OneAndTwoBitFlipsX{8}{1}{\OneBitFlipsTable}         \OneAndTwoBitFlipsX{8}{2}{\TwoBitFlipsTable}
        \OneAndTwoBitFlipsX{11}{1}{\OneBitFlipsTable}        \OneAndTwoBitFlipsX{11}{2}{\TwoBitFlipsTable}
        \OneAndTwoBitFlipsX{16}{1}{\OneBitFlipsTable}        \OneAndTwoBitFlipsX{16}{2}{\TwoBitFlipsTable}
        \OneAndTwoBitFlipsX{23}{1}{\OneBitFlipsTable}        \OneAndTwoBitFlipsX{23}{2}{\TwoBitFlipsTable}
        \OneAndTwoBitFlipsX{32}{1}{\OneBitFlipsTable}        \OneAndTwoBitFlipsX{32}{2}{\TwoBitFlipsTable}
        \OneAndTwoBitFlipsX{45}{1}{\OneBitFlipsTable}        \OneAndTwoBitFlipsX{45}{2}{\TwoBitFlipsTable}
        \OneAndTwoBitFlipsX{64}{1}{\OneBitFlipsTable}        \OneAndTwoBitFlipsX{64}{2}{\TwoBitFlipsTable}
    \end{semilogxaxis}
\end{tikzpicture}}

\newcommand{\LargeFlipMusings}{\begin{tikzpicture}
    \begin{semilogxaxis}[width=\linewidth, height=45ex, grid=both, legend pos=north west, legend columns=2, log base x=2, legend cell align={left}, xlabel={$\lambda$}, ylabel={Probability},
                         scaled ticks=false, y tick label style={ /pgf/number format/.cd, fixed, fixed zerofill, precision=2, /tikz/.cd }]
        \addplot coordinates { (64,0.02115604495485665) };
        \addlegendentry{\small $k=52$};
        \addplot coordinates { (64,0.004974873621168285) };
        \addlegendentry{\small $k=54$};
        \addplot coordinates { (64,0.0021456339262850704) };
        \addlegendentry{\small $k=55$};
        \addplot coordinates { (32,0.012670273367908326) };
        \addlegendentry{\small $k=57$};
        \addplot coordinates { (32,0.00527795292414326) };
        \addlegendentry{\small $k=59$};
        \addplot coordinates { (32,0.0033175898552511747) };
        \addlegendentry{\small $k=60$};
        \addplot coordinates { (16,0.0019078158633971584) };
        \addlegendentry{\small $k=61$};
        \addplot coordinates { (7,0.01100934480727602) (8,0.00463299559318884) (16,0.0019078141546296778)};
        \addlegendentry{\small $k=62$};
        \addplot coordinates { (16,0.008648380849678089) };
        \addlegendentry{\small $k=63$};
        \addplot coordinates { (6,0.009184070658384565) (7,0.008545949409381767) (8,0.001964886858224867) };
        \addlegendentry{\small $k=64$};
    \end{semilogxaxis}
\end{tikzpicture}}

\newcommand{\RuntimeComparisonsX}[3]{
    \addplot+ table[x=lambda,y expr={\thisrow{n}==#1 ? \thisrow{#2} - \thisrow{UUSD} : NaN}] {\RuntimeComparisonTable};
    \addlegendentry{\scriptsize #3};
}
\newcommand{\RuntimeComparisonsThree}{\begin{tikzpicture}
    \begin{loglogaxis}[width=\linewidth, height=33ex, grid=major,
                       legend style={at={(0.0,1.05)},anchor=south west}, legend columns=3, log base x=2, xlabel={$\lambda$}, ylabel={Runtime regret}, cycle list name=runtimecycle,
                       enlarge x limits=0.04, filter discard warning=false, unbounded coords=jump]
        \RuntimeComparisonsX{3}{RLS}{RLS}                                   \RuntimeComparisonsX{3}{k beta=1.3}{$\ell\sim\text{pow}(1.3)$} \RuntimeComparisonsX{3}{nevergrad beta=1.3}{$(1+\lambda)$ \emph{fast}, $\beta=1.3$}
        \RuntimeComparisonsX{3}{EA standard}{$(1+\lambda)\text{ EA}$}       \RuntimeComparisonsX{3}{k beta=1.5}{$\ell\sim\text{pow}(1.5)$} \RuntimeComparisonsX{3}{nevergrad beta=1.5}{$(1+\lambda)$ \emph{fast}, $\beta=1.5$}
        \RuntimeComparisonsX{3}{EA shift}{$(1+\lambda)\text{ EA}_{0\to1}$}  \RuntimeComparisonsX{3}{k beta=1.7}{$\ell\sim\text{pow}(1.7)$} \RuntimeComparisonsX{3}{nevergrad beta=1.7}{$(1+\lambda)$ \emph{fast}, $\beta=1.7$}
        \RuntimeComparisonsX{3}{EA resampling}{$(1+\lambda)\text{ EA}_{>0}$}
        \node[rectangle,draw,fill=yellow!40!white,anchor=south] at (2,1e-7) {$n=3$};
    \end{loglogaxis}
\end{tikzpicture}}
\newcommand{\RuntimeComparisonsHundred}{\begin{tikzpicture}
    \begin{loglogaxis}[width=\linewidth, height=33ex, grid=major,
                       legend pos=outer north east, log base x=2, xlabel={$\lambda$}, ylabel={Runtime regret}, every axis legend/.code={\renewcommand\addlegendentry[2][]{}}, cycle list name=runtimecycle,
                       enlarge x limits=0.04, filter discard warning=false, unbounded coords=jump]
        \RuntimeComparisonsX{100}{RLS}{RLS}                                   \RuntimeComparisonsX{100}{k beta=1.3}{$\ell\sim\text{pow}(1.3)$} \RuntimeComparisonsX{100}{nevergrad beta=1.3}{$(1+\lambda)$ \emph{fast}, $\beta=1.3$}
        \RuntimeComparisonsX{100}{EA standard}{$(1+\lambda)\text{ EA}$}       \RuntimeComparisonsX{100}{k beta=1.5}{$\ell\sim\text{pow}(1.5)$} \RuntimeComparisonsX{100}{nevergrad beta=1.5}{$(1+\lambda)$ \emph{fast}, $\beta=1.5$}
        \RuntimeComparisonsX{100}{EA shift}{$(1+\lambda)\text{ EA}_{0\to1}$}  \RuntimeComparisonsX{100}{k beta=1.7}{$\ell\sim\text{pow}(1.7)$} \RuntimeComparisonsX{100}{nevergrad beta=1.7}{$(1+\lambda)$ \emph{fast}, $\beta=1.7$}
        \RuntimeComparisonsX{100}{EA resampling}{$(1+\lambda)\text{ EA}_{>0}$}
        \node[rectangle,draw,fill=yellow!40!white,anchor=south] at (2,1e-2) {$n=100$};
    \end{loglogaxis}
\end{tikzpicture}}

\begin{document}

%%%%%%%%%%%%%%%%%%%%%%%%%%%%%%%%%%%%%%%%%%%%%%%%%%%%%%%%%%%
%%               Title, authors, abstract                %%
%%%%%%%%%%%%%%%%%%%%%%%%%%%%%%%%%%%%%%%%%%%%%%%%%%%%%%%%%%%

\title{Optimal Static Mutation Strength Distributions for the \texorpdfstring{$(1+\lambda)$}{(1+lambda)}~Evolutionary Algorithm on OneMax}
\thanks{This is a longer version of the paper accepted to GECCO'21 conference.}

\author{Maxim Buzdalov}
\affiliation{
\institution{ITMO University}
\city{Saint Petersburg}
\country{Russia}
}

\author{Carola Doerr}
\affiliation{
\institution{Sorbonne Universit{\'e}, CNRS, LIP6}
\city{Paris}
\country{France}
}

\begin{abstract}
Most evolutionary algorithms have parameters, which allow a great flexibility in controlling their behavior and adapting them to new problems. To achieve the best performance, it is often needed to control some of the parameters during optimization, which gave rise to various parameter control methods. In recent works, however, similar advantages have been shown, and even proven, for sampling parameter values from certain, often heavy-tailed, fixed distributions. This produced a family of algorithms currently known as ``fast evolution strategies'' and ``fast genetic algorithms''.

However, only little is known so far about the influence of these distributions on the performance of evolutionary algorithms, and about the relationships between (dynamic) parameter control and (static) parameter sampling. We contribute to the body of knowledge by presenting, for the first time, an algorithm that computes the optimal static distributions, which describe the mutation operator used in the well-known simple $(1+\lambda)$ evolutionary algorithm on a classic benchmark problem \textsc{OneMax}. We show that, for large enough population sizes, such optimal distributions may be surprisingly complicated and counter-intuitive. We investigate certain properties of these distributions, and also evaluate the performance regrets of the $(1+\lambda)$ evolutionary algorithm using commonly used mutation distributions.
\end{abstract}

%%%%%%%%%%%%%%%%%%%%%%%%%%%%%%%%%%%%%%%%%%%%%%%%%%%%%%%%%%%
%%                     ACM keywording                    %%
%%%%%%%%%%%%%%%%%%%%%%%%%%%%%%%%%%%%%%%%%%%%%%%%%%%%%%%%%%%

% \keywords{\revise{Mutation Operators, Runtime analysis}}%\carola{we can add this for camera-ready}

\maketitle

%%%%%%%%%%%%%%%%%%%%%%%%%%%%%%%%%%%%%%%%%%%%%%%%%%%%%%%%%%%
%%                     Discussion                       %%
%%%%%%%%%%%%%%%%%%%%%%%%%%%%%%%%%%%%%%%%%%%%%%%%%%%%%%%%%%%

%\section*{Discussion}
%\carola{just for our discussion:} 
%
%I just started writing, to have something on paper. Not super polished nor elegant, but at least a start..... So feel free to change whatever you want, of course. Also, some of this can possibly be merged into the intro -- let's see...
%
%Plots can be beautified once we know which ones we want to show. Either by you plotting in a different tool, or I could also do it in excel (and do proper export of the plots)

%%%%%%%%%%%%%%%%%%%%%%%%%%%%%%%%%%%%%%%%%%%%%%%%%%%%%%%%%%%
%%                     Introduction                      %%
%%%%%%%%%%%%%%%%%%%%%%%%%%%%%%%%%%%%%%%%%%%%%%%%%%%%%%%%%%%

\section{Introduction}
% What is commonly called an evolutionary algorithm is in most cases a family of different algorithms that share the same design patters. The algorithm instances can differ only in one parameter (e.g., ``the'' (1+1)~evolutionary algorithm (EA) can be instantiated with different mutation rates)  but they can also comprise algorithm instances that show significant diversity (e.g., several thousands of different instances of ``the'' CMA-ES~\cite{HansenO01} have been studied~\cite{modular-CMAES}).

Many evolutionary algorithms are composed of operators that are used within several algorithms. In the context of bit string representations, a very popular example of such an operator is \emph{standard bit mutation (SBM)}, the variation operator that takes as input a point $x$ (the ``parent''), modifies it by changing the entry in each position with some positive probability $p$ (independently of all other decisions), and outputs the so-called ``offspring'' $y$. SBM is a \emph{global} search operator, since the probability that it generates a particular point $y$ is positive, regardless of the input $x$. Another common mutation operator is the $\flipOp{1}$ operator, which creates the offspring by changing the entry in exactly one uniformly selected position. This search operator is a \emph{local} one. Both mutation operators are \emph{unbiased} in the sense proposed by Lehre and Witt~\cite{LehreW12}, i.e., they treat all positions and all possible values identically.

By a characterization derived in~\cite{DoerrDY20}, unary unbiased operators are exactly the ones that can be described by a distribution $D$ over the integers $\{0, 1, \ldots, n\}$. To apply them, one first samples a \emph{mutation strength} $\ell$ from this distribution and then changes the input by applying the $\flipOp{\ell}$ operator, which changes the entry in $\ell$ uniformly chosen, pairwise different positions. SBM can be exactly characterized by the binomial distribution $\Bin(n,p)$ with $n$ trials (one for each position) and success probability $p$ (success=bit flip), hence it is unbiased. Likewise, $\flipOp{1}$ is the operator associated with the 1-point distribution assigning all probability mass to mutation strength 1.

% By identifying mutation operators with their distributions over the possible mutation strengths, we can easily interpolate between them -- an effect that can be very beneficial when aiming at a smooth convergence from a global to a local search behavior.
%The characterization provided in~\cite{DoerrDY20} also allows to define new mutation operators by simply defining a distribution over the possible \emph{mutation strengths} $0 \le k \le n$. 
% This view also invites to define new mutation operators, an avenue that was very successfully taken by the ``fast'' mutation operator introduced in~\cite{fastGA}. 

\textbf{Our contribution:} 
We investigate in this work how common mutation operators, such as the ones mentioned above or the fast mutation operator suggested in~\cite{fastGA}, compare to an optimal one. 
To analyze this question, we introduce the $(1+\lambda)$~UUSD-EA, the family of all $(1+\lambda)$-type mutation-only algorithms whose mutation operator can be defined via a \underline{u}nary \underline{u}nbiased \underline{s}tatic \underline{d}istribution (which may depend on the problem dimension $n$ and the offspring population size $\lambda$, but may not change during the run). This family comprises all $(1+\lambda)$~evolutionary algorithms (EAs), their Randomized Local Search (RLS) counterparts (which use the $\flipOp{1}$ operator instead of SBM), the fastGAs, the normalized EAs~\cite{YeDB19}, etc. 

For various combinations of $n$ and $\lambda$ we numerically compute the minimal expected runtime that can be obtained by any $(1+\lambda)$~UUSD-EA on \onemax in dimension $n$, and we compare these runtimes to that of the classically studied $(1+\lambda)$-type algorithms mentioned above. Our lower bounds are constructive, in that we also derive the distributions associated with these optimal $(1+\lambda)$~UUSD-EAs. This allows us to study the properties of these optimal distributions.

\textbf{Approach:} Since the optimal distributions cannot be determined by exact analytical approaches, we apply the separable CMA-ES~\cite{sep-cma-es} to identify them. It shows very good performance, and provides us with distributions that are optimal up to the last few digits in the available machine precision. Surprisingly, the separable version of CMA-ES is not only computationally faster, but also yields results of the same or better quality than more sophisticated versions of CMA-ES.

\textbf{Main result:} Our numerical results show, among other things, that the $\flipOp{1}$ operator is optimal when $\lambda$ is small, whereas the conditional binomial distribution $\Bin_{>0}(n,1/2)$ is nearly optimal when $\lambda$ is very large: with this distribution, the $(1+\lambda)$~UUSD-EA performs a uniform random search. 
The optimal distributions in the middle regime show a rather complex behavior, for which we can identify a few patterns, but we also observe a few phenomena that may look counter-intuitive at first glance.
For instance, the first mutation strength different from one-bit flip that gets nonzero probability when $\lambda$ grows appears to be either $n$ or $n-1$.

\textbf{Relationship to black-box complexity and to parameter control:} Our work can be seen as a continuation of black-box complexity theory for $\lambda$-parallel~\cite{ParallelBBC14} elitist~\cite{DoerrL17ECJ} unary unbiased~\cite{LehreW12} black-box algorithms with static configuration. A key advantage of lower bounds such as ours is that it allows to rigorously quantify the impact of individual decisions. In our case, the driving motivation behind our analyses is a rigorous quantification of the difference between static and dynamic algorithm configurations. Put differently, we aim at quantifying the gap between the algorithms using parameter control and the ones that do not. In contrast to classical runtime and black-box complexity results, our work focuses on an exact numeric evaluation of this gap for concrete problem dimensions.  
% \maximm{There can be gaps between lower bounds on the black-box complexity with respect to some constraint on employed algorithms (e.g. $\Omega(n \log n)$ for unary unbiased algorithms) and upper bounds on the black-box complexity without that constraint, or with fewer restrictions (e.g. $O(n)$ for binary unbiased algorithms). Such gaps are equivalent to statements that a certain feature (e.g. having crossovers) is necessary in an algorithm so that the problem is solved efficiently enough. Our work builds a foundation for answering similar questions about parameter control: we essentially produce numerically precise lower bounds on what is possible without the active forms of parameter control that employ non-trivial dependencies on the optimization history, or at least on the index of the offspring being generated.} % Strictly speaking, 2-rate also does the latter.
%\carola{To be done. Lower bound already available from unbiased black-box complexity. Our work is different in that (1) we want to derive \emph{upper} bounds by determining the optimal distributions and (2) we aim at a higher level of precision, to quantify the impact}

\textbf{Impact:} While the results of our concrete analysis may mostly appeal to theoreticians, our work invites to take a different view on mutation operators by defining them via distributions over the possible mutation strengths. This alternative view makes it substantially easier to generalize concepts such as SBM, $\flipOp{1}$, etc.~-- an advantage that can be leveraged, for example, for designing smooth convergence from global to local search behavior. But the design principle can also lead to performance gains in the static setting. A first strong result in this context is the fastGA~\cite{fastGA}, which has become the new state-of-the-art in several applications~\cite{NGopt}. Our work indicates that there is quite some untapped potential in the design of new mutation operators, and we hope that our work inspires new work in this direction. 

Based on the findings of our work, we formulate two conjectures, for which we do not yet have formal proofs.

\textbf{Conjecture 1:} For each $n$ there exists a threshold $\lambda_1(n)$ such that for all $\lambda \le \lambda_1(n)$ the 1-point distribution is optimal. Our guess on the particular dependency is $\lambda_1(n) = \Theta(\log n)$.

\textbf{Conjecture 2:} For each $n$ and each arbitrarily small $\varepsilon$ there exists another threshold $\lambda_b(n, \varepsilon)$ such that, for all $\lambda \ge \lambda_b(n, \varepsilon)$, the optimal distribution is closer than $\varepsilon$ in any reasonable metric~-- such as the maximum of differences between probabilities across all mutation strengths~-- to the conditional binomial distribution $\Bin_{>0}(n,1/2)$. This is equivalent to stating that the uniform random search is arbitrarily close to being optimal for large enough $\lambda$.
%\carola{Do we want to state this?} We conjecture that for each $n$ there exists a threshold $\lambda(n)$ such that for all $\lambda>\lambda(n)$ the binomial distribution $\Bin(n,1/2)$ is optimal. This conjecture is equivalent to stating that uniform random search is optimal for large enough $\lambda$. We currently do not have a formal proof for this.

\textbf{Related work:} Our study continues our recent works~\cite{BuskulicD19} and~\cite{BuzdalovD20}, which provide optimal dynamic configurations for $(1+1)$ and $(1+\lambda)$-type algorithms, respectively. While their works are restricted to specific mutation operators (variants of SBM and $\flipOp{k}$), we study in this work a generalization to arbitrary unary unbiased variation operators. In contrast to~\cite{BuskulicD19,BuzdalovD20} we focus on \textit{static} configurations, with the idea to build a rigorous baseline against which we can compare dynamic parameter control methods. 

For $\lambda=1$, the work~\cite{DoerrDY20} quantifies the asymptotic advantage of the best unary unbiased algorithm with dynamic distributions against the best static one (RLS). To extend this work to $(1+\lambda)$-type algorithms, a rigorous bound on the best static case is needed~-- a baseline that we provide in this work for various combinations of $n$ and $\lambda$, with the hope that the insights generated by our examples can be leveraged to rigorously prove certain characteristics of the optimal static unary unbiased operators. 
% A baseline for the latter was provided in~\cite{DoerrDY20}, where it is shown that the  the expected runtime of the drift-maximizer and that of the optimal unary unbiased $(1+1)$-type algorithm cannot be larger than $\Theta(n^{2/3}\log^{9} n)$.

For $\lambda>1$, related works can be found in the context of the parallel black-box complexity model~\cite{ParallelBBC14,ParallelBBC15,LehreS20arxivparallelBBC} and for the \opl~\cite{GiessenW17,GiessenW18}. All these works, however, are either less interested in exact runtime bounds (and focus on asymptotic runtime guarantees instead) or they concern specific mutation operators only. 
For generalized mutation operators, concrete examples can be found in the mentioned works~\cite{fastGA,YeDB19}. We are not aware, however, of previous works explicitly studying optimal unary mutation operators. 

\textbf{Availability of code and data:} All project source code and data are available for public use at~\cite{data}.

%%%%%%%%%%%%%%%%%%%%%%%%%%%%%%%%%%%%%%%%%%%%%%%%%%%%%%%%%%%
%%                     Preliminaries                     %%
%%%%%%%%%%%%%%%%%%%%%%%%%%%%%%%%%%%%%%%%%%%%%%%%%%%%%%%%%%%
\section{From Mutation Operators to Mutation Strength Distributions}
\label{sec:preliminaries}

%We use the notation $[a..b]$ to denote a set of integers that fall into the real interval $[a,b]$,
%and we abbreviate $[1..n]$ by $[n]$.

% \subsection{Unary Unbiased Variation}
% \label{sec:unary} 

We are concerned in this work with a generalizing view on unary unbiased variation operators, often referred to in the evolutionary computation context as \emph{``mutation''}. In a nutshell, a mutation operator takes as input a search point $x \in \mathcal{S}$, $\mathcal{S}$ denoting the search space, and creates from it an offspring $y \in \mathcal{S}$. More formally, a mutation operator is a family $(D(x))_{x \in \mathcal{S}}$  of unary distributions over the search space $\mathcal{S}$. When fed with an input $x$, a new search point $y$ is sampled from $D(x)$. 

One of the most common mutation operators is standard bit mutation (SBM). 
In the context of pseudo-Boolean optimization (i.e., the maximization of a function $f:\{0,1\}^n \to \R$)~-- which is the setting that we assume for the remainder of the paper~-- SBM is often explained as follows: to obtain an offspring $y$ from $x$, we first create a copy of $x$ and then we decide for each bit position $i \in [n]:=\{1,\ldots,n\}$ whether the entry shall be updated to $1-x_i$ (``bit flip'') or whether the current entry is maintained.%\footnote{For two real numbers $a$ and $b$, we denote by $[a..b]$ the set of integers that fall into the interval $[a,b]$. We abbreviate $[1..n]$ by $[n]$.}
The bit flip decisions are made independently of each other. The probability $p \in (0,1]$ to flip an entry is referred to as the \emph{mutation rate}.

SBM is a prime example of a \emph{unary unbiased mutation operator} in the sense proposed by Lehre and Witt in~\cite{LehreW12}. By a characterization proven in~\cite[Lemma~1]{DoerrDY20}, this class subsumes all variation operators that are fully described by a distribution $D$ over the possible mutation strengths $\ell \in [0..n]:=[n] \cup \{0\}$. When applied to a search point~$x$, the operator first samples a mutation strength $\ell \in [0..n]$ from its operator-specific distribution~$D$ and then creates the offspring $y$ by flipping the entries in $\ell$ pairwise different, uniformly selected entries. 

It is not difficult to see that the operator-specific distribution of SBM is the binomial distribution $\Bin(n,p)$ with $n$ trials and success probability $p$. Another common operator is the 1-bit-flip operator $\flipOp{1}$, which is used, for example, within Randomized Local Search (RLS). $\flipOp{1}$ creates the offspring by flipping exactly one uniformly chosen bit. Its operator-specific distribution over the mutation strengths $[0..n]$ is hence the 1-point distribution that assigns all probability mass to $\ell=1$. Likewise, the $\flipOp{k}$ operator flips $k$ pairwise different, uniformly selected bits, and its operator-specific distribution is the 1-point distribution giving all probability mass to $\ell=k$. 
Other common unary unbiased mutation operators include the ``shift'' SBM, $\SBM_{\shift}$, which is similar to SBM but which assigns all probability weight from $\ell=0$ to $\ell=1$, and the ``resampling'' SBM, $\SBM_{>0}$, which assigns the probability weight of sampling mutation strength $0$ proportionally to all positive mutation strengths $1 \le \ell \le n$ by assigning to each of these values probability $\Bin(n,p)(\ell)/(1-(1-p)^n)$; see~\cite{CarvalhoD18arxiv,CarvalhoD18PPSN} for a discussion and motivation of these two latter variants. 
Motivated by the observation that infrequent large ``jumps'' can be beneficial in evolutionary algorithm behavior, B.~Doerr et al. introduced in~\cite{fastGA} the \emph{fast genetic algorithm (GA)}, which samples the mutation strength from the heavy-tailed, power-law distribution $\mathcal{P}[\ell=k]=(C_{n/2}^{\beta})^{-1}k^{-\beta}$ with $C_{n/2}^{\beta}=\sum_{i=1}^{n/2} i^{-\beta}$ and $\beta$ being some constant, often set as $\beta=1.5$. 
Finally, in~\cite{YeDB19} a normalized mutation operator was suggested, which samples the mutation strength from a normal distribution $\mathcal{N}(\mu,\sigma^2)$. In contrast to the examples discussed above, this operators allows to scale the mean and the variance of the distribution independently of each other. 

The characterization from~\cite[Lemma~1]{DoerrDY20}, which identifies mutation operators via their distributions over the set $[0..n]$, is classically only used to verify that a certain operator is unbiased. We use it here the other way around, by asking ourselves how different the optimal mutation operators are from those that are commonly used in evolutionary computation.

%inspires an interesting generalization of the classical mutation operators, which can from now on be defined by any distribution over the set $[0..n]$. The main purpose of this paper is to demonstrate that this alternative view on mutation operators can lead to very surprising findings. 
% In fact, we note that even for well-researched problems such as \onemax, we currently do not know what the optimal sampling distributions looks like, and this not even when we restrict our attention to \emph{static distributions}, as we shall do in the remainder of this paper.    

% \subsection{The \oplheader~UUSD-EA}
\textbf{The \oplheader~UUSD-EA.} 
We study this question in the context of the $(1+\lambda)$~unary unbiased static distribution EA (UUSD-EA), which is given by Algorithm~\ref{alg:opl}. The $(1+\lambda)$~UUSD-EA is initialized uniformly at random. In each iteration, it samples $\lambda$ points, which are all sampled from the same unary unbiased mutation operator. We denote the distribution from which the mutation strength is sampled by $D(n, \lambda)$ to indicate that it may depend on $n$ and $\lambda$, but not on any information accumulated during the run of the algorithm. That is, the $(1+\lambda)$~UUSD-EA allows only \emph{static} mutation operators. For creating the $\lambda$ search points that shall be evaluated in the current iteration, the $(1+\lambda)$~UUSD-EA samples for each one of them a mutation strength $\ell(i)$ and then creates the $i$-th ``offspring'' by applying the $\flipOp{\ell(i)}$ operator, which flips $\ell(i)$ pairwise different, uniformly chosen bits in the input $x$. The best of these offspring replaces $x$ if it is at least as good as it. %We can assume ties to be broken uniformly at random, but the results presented in this paper apply to any other tie-breaking rule. 
It is irrelevant for the context of our work how the ties are broken in line~\ref{line:select}, as our results apply to all tie-breaking rules.  

\begin{algorithm2e}[t]%
\caption{The $(1+\lambda)$ unary unbiased static distribution EA (UUSD-EA) maximizing a function $f:\{0,1\}^n \rightarrow \R$.}\label{alg:opl}
%Operators are defined in the main text.
	\textbf{Initialization:} 
	Sample $x \in \{0,1\}^{n}$ uniformly at random and evaluate $f(x)$\;
    \textbf{Optimization:}
    \For{$t=1,2,3,\ldots$}{
		\For{$i=1,\ldots,\lambda$}{
			\label{line:k}
			Sample $\ell(i) \sim D(n, \lambda)$\;
			\label{line:mut} $y^{(i)} \leftarrow \flipOp{\ell(i)}(x)$\;
			Evaluate $f(y^{(i)})$\;
		}%for i
		$y \leftarrow \select\left(\arg\max\{f(y^{(i)}) \mid i \in [\lambda]\} \right)$\; 	
		\lIf{$f(y)\geq f(x)$}{$x \leftarrow y$\label{line:select}}	
	}%for t
\end{algorithm2e}

\textbf{\om:} We focus on \onemax, i.e., our goal is to determine the optimal static unary unbiased distributions for maximizing the function $\sum_{i=1}^n{x_i}$. In the context of our work, this problem is equivalent to that of minimizing the Hamming distance to an arbitrary bit string $z \in \{0,1\}^n$~\cite{LehreW12}. 

\textbf{Expected runtimes:} As common in literature on theory of evolutionary computation, we understand as an optimal distribution the one that minimizes the expected runtime, which we measure here in terms of generations. Since the offspring population size $\lambda$ is fixed, this does not impact our results: a parallel runtime of $T$ generations corresponds to a runtime of exactly $(T-1)\lambda+1$ function evaluations. As \emph{runtime} we therefore denote the number of iterations that the algorithm performs until it evaluates an optimal solution for the first time.  

\textbf{(Non-)uniqueness of the optimal distributions:} We note that we do not have any guarantee at the moment that the optimal distributions are unique. In fact, we observe that for any fixed problem dimension $n$, there is a certain threshold $\lambda(n)$ before which small differences in the distributions cause measurable effects on the expected running times, so that the optimal distribution seems to be unique from the point of view of our computations. After the threshold, however, differences between the distributions have no measurable effect on the expected runtime, so that our algorithm may consider different ones as optimal, and may hence not always return the same distribution.

\textbf{Notation:} For combinations of $n$ and $\lambda$ for which the optimal distributions are unique, we denote by $P^*(k \mid n, \lambda)$ the probability that this distribution assigns to flipping $k$ bits, $k \in [n]$. To ease the reading, we sometimes use the same notation also for those distributions which yield expected running times that deviates only negligibly from what appears to be the true optimum.
 
%%%%%%%%%%%%%%%%%%%%%%%%%%%%%%%%%%%%%%%%%%%%%%%%%%%%%%%%%%%
%%                     The Algorithm                     %%
%%%%%%%%%%%%%%%%%%%%%%%%%%%%%%%%%%%%%%%%%%%%%%%%%%%%%%%%%%%

\section{Algorithm for Computing the Optimal Distributions}
\label{sec:algo}

Our algorithm to compute the optimal static unary unbiased distributions is based on the dynamic programming approach from~\cite{BuzdalovD20}. We start the description by explaining the principles of dynamic programming that are used to compute the expected running time $T_f$ of the $(1+\lambda)$~UUSD-EA, measured in iterations, assuming it starts at fitness $f$ and the values $T_{f'}$ are known for all $f' > f$. 
We note then that a (practical) analytic solution of the problem of finding an optimal distribution is quite unlikely to exist even for small values of $n$, and instead give a black-box minimization scheme with the use of a separable CMA-ES~\cite{sep-cma-es}, a simplified and more computationally efficient version of the well-known continuous optimizer~\cite{HansenO01}. We complete with an investigation of convergence properties, which allows us to say, with great confidence, that separable CMA-ES finds a globally optimal distribution in a constant fraction of runs.

\subsection{Dynamic Programming on Expected Times}

% \carola{all of this has been said above, I do not think we need to repeat it here} We consider the $(1+\lambda)$~UUSD-EA with population size $\lambda$ on \onemax with problem size $n$, whose definition $D$ is explicitly described by the probabilities $(D_k)_{k = 0,1,\ldots,n}$ of flipping exactly $k$ bits. Assume further that its current state is unambiguously described by the parent fitness $f$, and for all higher fitness values $g > f$ the expected times to reach the optimum $T_g$ are computed (where it trivially holds that $T_n = 0$). The goal of this step is to compute $T_f$.

We first explain how to compute $T_f$ for a given distribution $D=(D_k)_{k \in [0..n]}$. 
We begin with computing the probabilities $S_{k,g}$ of sampling an offspring with fitness $g$ by flipping exactly $k$ bits chosen uniformly without replacement in a solution of fitness $f$. Note that these quantities depend only on the current fitness $f$ and the problem properties, that is, they depend neither on $\lambda$ nor on $D$. It holds from simple combinatorics that $S_{k,g} = \binom{n-f}{g-f+i} \binom{f}{i} / \binom{n}{k}$, where we assume this probability to be zero if one of the binomial coefficient arguments are out of bounds.

Next we compute the probabilities $(Q^{(1)}_g)_{g = f+1,f+2,\ldots,n}$ of sampling a \emph{single} offspring of fitness $g$. For $g > f$ they are derived from $S_{k,g}$ by using the distribution parameters $(D_k)_{k \in [0..n]}$ as follows:
\begin{equation*}
    Q^{(1)}_g = \sum\nolimits_{k = 1}^{n} D_k S_{k,g}.
\end{equation*}

As the $(1+\lambda)$~UUSD-EA is an elitist algorithm, and the behavior of the algorithm with different parents having the same fitness value is the same, with the remaining probability the algorithm remains in the same state, which we capture as $Q^{(1)}_f = 1 - \sum_{g = f+1, f+2, \ldots, n} Q^{(1)}_g$.

The probability of each possible fitness improvement after sampling all the $\lambda$ offspring is then computed as follows:
\begin{equation}
    Q^{(\lambda)}_g = \left(\sum\nolimits_{i=f}^{g} Q^{(1)}_i\right)^{\lambda} - \left(\sum\nolimits_{i=f}^{g-1} Q^{(1)}_i\right)^{\lambda}. \label{eq:transitions-lambda}
\end{equation}
In simple words, the new fitness is $g$ if all offspring have the fitness in $[f..g]$, but not all of them have the fitness in $[f..g-1]$, counting all offspring with fitness smaller than $f$ towards $f$.

Finally, the expected time to reach the optimum from the fitness $f$ is computed using the following expression:
\begin{equation}
    T_f = \frac{1}{1 - Q^{(\lambda)}_f} \left(1 + \sum\nolimits_{g = f+1}^{n} Q^{(\lambda)}_g T_g\right), \label{eq:tf}
\end{equation}
% the derivation of which follows the standard techniques.
by standard arguments as detailed in~\cite{BuskulicD19,BuzdalovD20}.

The time and memory complexities of such a step is $O(n^2)$ for computing $S_{k,g}$ and $O(n)$ for the other stages. As there are $n$ different fitness values for $f$, the time complexity of computing all the expected running times for one set of algorithm parameters $(\lambda, D)$ is $O(n^3)$, whereas the memory complexity is still $O(n^2)$ as the $S_{k,g}$ matrix may be discarded once $f$ changes. However, since $S_{k,g}$ depend only on $n$ and $f$, one may evaluate up to $O(n)$ different combinations of algorithm parameters $(\lambda, D)$ in a single run by merging the activities corresponding to each $f$, which preserves the total time complexity of $O(n^3)$ and the memory complexity of $O(n^2)$. This feature will turn later to be beneficial when a population-based optimizer is applied to find the best possible distribution $D$.

\subsection{Optimization with Separable CMA-ES}

When solving the problem of finding an optimal distribution $D$, one could choose to express each of $T_f$ as a function of $n+1$ distribution parameters $(D_k)_{k \in [0..n]}$, promote such expressions through dynamic programming, take their weighted sum for the expected running time from a random point and perform analytical optimization using standard analysis approaches. However, the presence of $Q^{(\lambda)}_f$ in the denominator in~\eqref{eq:tf} makes the resulting expression nonlinear even for $\lambda=1$, and having a $\lambda$ in the exponent in~\eqref{eq:transitions-lambda} makes it even harder. The resulting expression appears to be a ratio of polynomials of degree $\Theta(n\lambda)$ with $D_k$ as variables, which makes it infeasible to perform an exact analytical minimization even for small problem sizes.

Such large degree of the polynomials also effectively prevents gradient-based optimization, since the exact numeric computation of the derivatives~-- although possible~-- would require considerable computational resources, which furthermore significantly increase with $\lambda$. At the same time, the time required to evaluate the expected runtime of the $(1+\lambda)$~UUSD-EA for a given distribution does not depend on $\lambda$, assuming the computations are done in machine precision. This makes it possible to apply black-box optimization techniques to identify distributions which minimize the expected runtime. From the large set of possible black-box optimization techniques, we chose the CMA-ES family of algorithms~\cite{HansenO01} that are well suitable for such kind of optimization. The application of CMA-ES requires a few clarifications, which we list below.
\begin{itemize}
    \item Our implementation of CMA-ES is based on the one from Apache Commons Math, version 3.6.1. The choice of the Java programming platform was due to the computational complexity of the fitness evaluation. On one hand, it is too expensive to be implemented in Python or Matlab (the languages with reference implementations of CMA-ES maintained by the authors of this algorithm). On the other hand, the costs of inter-process communication are not negligible compared to fitness evaluation, so evaluation of fitness shall happen in the same process which runs the optimizer.
    \item To search the distributions, we tune the CMA-ES to respect box constraints (each variable is in $[0,1]$) and, before fitness evaluation, we normalize variable values without updating the individual, which would clash with the assumptions made by CMA-ES.
    \item As we perform distribution optimization for a noiseless problem, we can safely set $D_0$ to zero, which restricts the search to the $n$-dimensional cube $[0,1]^n$.
\end{itemize}

We had to modify the implementation of CMA-ES, as the particular class hierarchy in Apache Commons Math does not allow to evaluate the whole population in a single call, which we benefit from. Besides, after some experimentation, we switched to the separable CMA-ES~\cite{sep-cma-es}, which is more efficient in terms of computational costs, but also produced better results (likely due to fewer operations that caused precision loss). We also vastly optimized its implementation, which resulted in an $8\times$ reduction of the wall-clock running time.

The configurable parameters of CMA-ES were: population size $10$, initial step size $1.0$, random initial guess, computational budget $100 n^2$. The optimizer, however, did not reach the computational budget, as, in all runs, it converged to a single point and terminates at one of the degeneration criteria much earlier than that.

\subsection{Convergence Analysis}

\begin{figure}[!t]
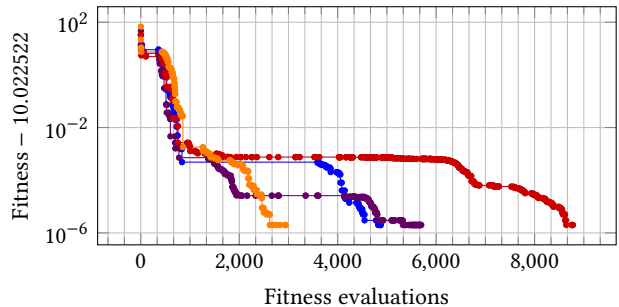

\ConvergenceCurves
\caption{Fitness as a function of the number of evaluations for $n=16$ and $\lambda=8$. First four runs are shown.}\label{fig:conv:curves}
\end{figure}

\begin{figure}[!t]
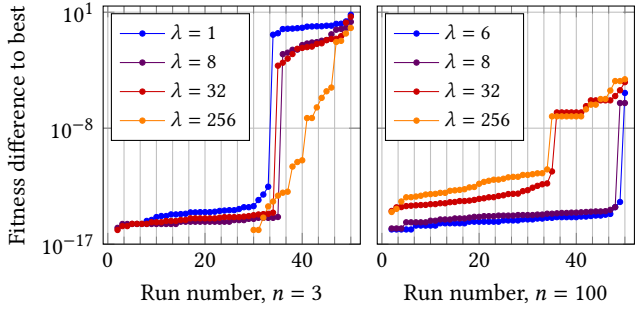

\ConvergencePrecisionThree\ConvergencePrecisionHundred
\caption{Relative loss of the expected runtime induced by the result of the 50 CMA-ES runs, when compared against the best expected runtime. Plots are for $n=3$ (left) and $n=100$ (right), respectively, values are sorted by increasing loss.  
%s of best distributions found in 50 runs for $n\in\{3,100\}$,
}
\label{fig:conv:time-precision}
\end{figure}

The preliminary experiments showed that CMA-ES typically converges relatively quickly to nearly the same value in most of runs. Fig.~\ref{fig:conv:curves} shows example runs
for $n=16$ and $\lambda=8$. Runs for all $n$ and $\lambda$ demonstrate similar behavior. The rest of the paper is based on the data collected for
$n \in \{ 3, 5, 8, 11, 16, 23, 32, 45, 64, 91, 100 \}$ and $\lambda \in [1..7] \cup \{ 2^i \mid 3 \le i \le 10 \}$. For each $(n, \lambda)$ pair we performed 50 independent runs of the CMA-ES.

We observed that in a constant fraction of runs the algorithm optimized the distribution to produce the same expected running time up to precision of $10^{-12}$ and better.
Fig.~\ref{fig:conv:time-precision} shows examples for some values of $\lambda$ for $n=3$ and $n=100$. Note that there is no curve for $\lambda=1$ in the $n=100$ plot since in this case all
the values were identical. Since the delivered precision is very close to the precision available for the 64-bit floating point machine numbers, we assume that CMA-ES reaches the global optimum of the problem in most of the cases. For further analyses, we selected only the runs whose result exceeds the obtained minimum by a factor of at most $(1 + 10^{-9})$.

\begin{figure}[!t]
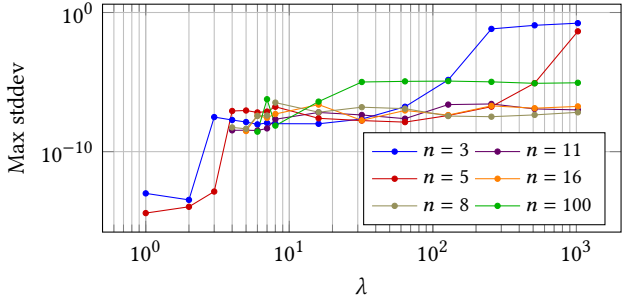

\ConvergenceInDistributions
\caption{Maximal standard deviations, out of $k$ standard deviations of $P^*(k \mid n, \lambda)$, as functions of $\lambda$ for various $n$.}\label{fig:conv:dist-std}
\end{figure}

We have also performed the robustness analysis for the distributions produced by the optimizer. For each $n$, $\lambda$, and $k$, the standard deviation of the values $P^*(k \mid n, \lambda)$ was computed across all the good enough distributions. The results are presented in Fig.~\ref{fig:conv:dist-std}, where an intriguing picture appears. First, small $\lambda$ produce very small (much less than $10^{-10}$) maximal standard deviations, which then jump to the region of $[10^{-8}; 10^{-5}]$ and remain there until $\lambda$ reaches a certain threshold. Above that threshold, the maximal standard deviations experience some sort of phase shift and raise to very high values reaching $0.1$ and above. Note that, by our selection of the data that is considered in this computation, the $(1+\lambda)$ UUSD-EA still shows nearly identical expected running times on such different distributions and values of $\lambda$. We show later in the next section that this is, in fact, an expected behavior that corresponds to situations when there is a single global optimum, but a number of different distributions coincide in its running time expectation with that global optimum up to the machine precision.

\begin{figure}[!t]
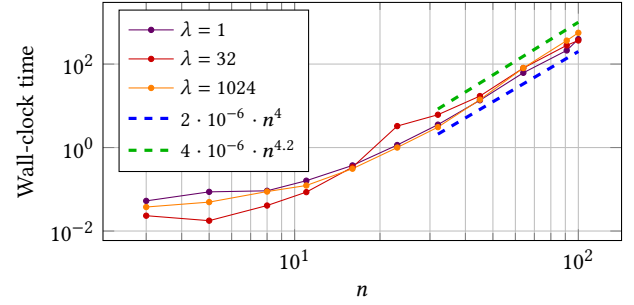

\WallClockTimes
\caption{Average wall-clock times for various $\lambda$ together with guesses for asymptotic bounds.}\label{fig:conv:wallclock}
\end{figure}

Fig.~\ref{fig:conv:wallclock} displays the average wall-clock times required to find the optimal distribution for all available $n$ and few values of $\lambda$.
The available data suggests that the time complexity scales polynomially with the degree of $4+\varepsilon$ for some small $\varepsilon$, which, together with the earlier cubic
runtime bound for the evaluation of a given distribution, suggests that the CMA-ES requires $O(n^{1+\varepsilon})$ iterations before hitting one of its termination criteria. 

%%%%%%%%%%%%%%%%%%%%%%%%%%%%%%%%%%%%%%%%%%%%%%%%%%%%%%%%%%%
%%                   Distributions                       %%
%%%%%%%%%%%%%%%%%%%%%%%%%%%%%%%%%%%%%%%%%%%%%%%%%%%%%%%%%%%

\section{Optimal Distributions}
\label{sec:distributions}

We now take a closer look at the distributions returned by the algorithm from Sec.~\ref{sec:algo}. As mentioned, the data is available at~\cite{data}, and we present here only our main findings. 

% MB: Removed this title as the second subsection is nearly empty compared to this one, and I commented that one out. 
%\subsection{Impact of \texorpdfstring{$\lambda$}{lambda} on \texorpdfstring{$P^*(k \mid n, \lambda)$}{Optimal Distribution Parameters} for Fixed \texorpdfstring{$n$}{N}}

\begin{figure}[!t]
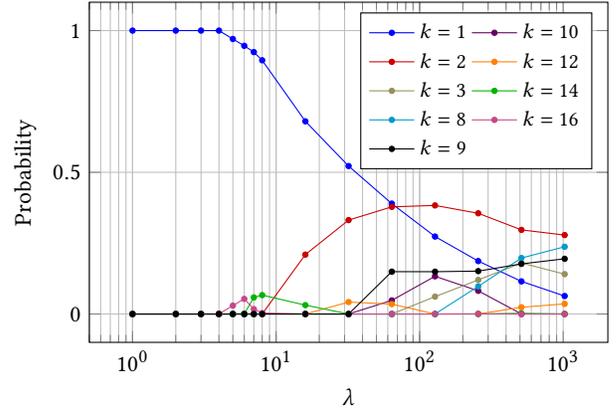

\DistributionsByLambdasSixteen
\caption{Probabilities of flipping $k$ bits in optimal static mutation strength distributions for different population sizes $\lambda$ for optimizing \om in dimension $n=16$.}
\label{fig:16-lines}
\end{figure}

We first study the impact of $\lambda$ on the optimal distributions. To this end, we fix the dimension and analyze how for each possible mutation strength $k \in [n]$ its probability of being chosen depends on $\lambda$. Fig.~\ref{fig:16-lines} illustrates these probabilities for dimension $n=16$. 

% A few general trends, which can be observed also in all larger dimensions \carola{we need to carefully check this!} are as follows. 

\textbf{The 1-point distribution $\bm{\Pr[\ell=1]=1}$ is optimal when $\bm{\lambda}$ is small.} For $n=16$, we see that deterministically flipping one bit ($\Pr[\ell=1]=1$) is optimal for $\lambda \le 4$. Note that for $\lambda=1$ this distribution defines Randomized Local Search (RLS), an algorithm that is often used as baseline for comparisons, both in empirical~\cite{DoerrYHWSB20} and in theoretical~\cite{DoerrN20} research. %RLS is also well studied in the theory of evolutionary computation literature, since it often provides general proof ideas that can then be extended to more complex evolutionary algorithms~\carola{would we have a good reference? If not, we can cite~\cite{DoerrN20}}. 
Our data shows that for $n\in\{8,11\}$ the generalized $(1+\lambda)$~RLS is optimal for $\lambda \le 3$ and suboptimal for $\lambda=4$. 
Similarly, for $n \in \{ 16,23,32 \}$ it is optimal for $\lambda \le 4$, and for $n \in \{ 64, 91, 100\}$ the threshold is $\lambda = 5$. 
We are confident that this describes a general trend of a positive correlation between the dimension $n$ and the maximal $\lambda$ for which the 1-point distribution $\Pr[\ell=1]=1$ is optimal.

\begin{figure}[!t]
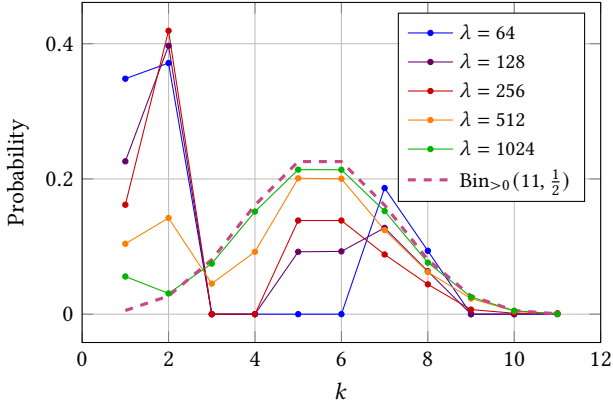

\DistributionsByDistanceEleven
\caption{Values $P^*(k \mid n=11, \lambda)$ for large $\lambda$ plotted against $k$. The larger $\lambda$, the more the curves resemble the binomial distribution $\Bin_{>0}(11,1/2)$, which is also plotted.}
\label{fig:n11-large}
\end{figure}

\begin{figure}[!t]
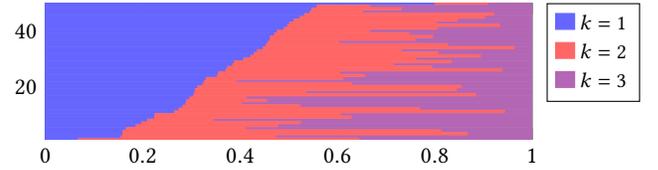

\DistributionsThreeNoise
\caption{$P^*(k \mid n=3, \lambda=1\,024)$ for each of the 50 independent runs of our optimizer. The expected optimization time is identical for all of them (0.875).}
\label{fig:n3-L1024}
\end{figure}

\textbf{The conditional binomial distribution $\Bin_{>0}\bm{(n,1/2)}$ is arbitrarily close to an optimal one for large $\bm{\lambda}$.} 
We plot in Fig.~\ref{fig:n11-large} the optimal probability $P^*(k \mid n=11, \lambda)$ for large $\lambda$. The curves for $\lambda=512$ and $\lambda=1\,024$ approximate the conditional binomial distribution $\Bin_{>0}(11,1/2)$, which we plot as dotted red line. This can be explained as follows: the $(1+\lambda)$~UUSD-EA with the static mutation strength distribution $\Bin_{>0}(n,1/2)$ is simply the random search algorithm, which samples all search points, except the parent, uniformly at random. For $\lambda = \Omega(2^n)$, this algorithm has a very good chance of sampling every point $y \in \{0,1\}^n$, so that it also has a decent chance of hitting the optimum. When $\lambda$ is not much larger than $2^n$, the quality of a distribution is significantly influenced by each of its parameters, so for these cases our computations provide distributions that are almost identical to random search in each independent run. 
When $\lambda$ is much bigger than $2^n$, however, the situation changes: while the common sense suggests that the truly optimal distribution gets even closer to pure random search, the quality of a distribution quickly loses sensitivity to its parameters, and we obtain different distributions, which all yield practically indistinguishable expected runtimes.
We provide an example for $n=3$ in Tab.~\ref{tab:n3}, where we observe that for $\lambda \le 128$ the average of the computed static mutation strength distributions converge against $\Bin_{>0}(n,1/2)$. The standard deviation of the independent runs of our optimizer are negligible in this regime. For $\lambda=256$, $512$, and $1\,024$, however, the maximum standard deviation among the three mutation strengths are 0.07, 0.12, and 0.17, respectively. 
Fig.~\ref{fig:n3-L1024} plots the results of all 50 independent runs for $\lambda=1\,024$.

\begin{table}[!t]
    \caption{Average values of the recommended distributions $P^*(k \mid n=3, \lambda)$ for different $\lambda$ and all mutation strengths $k$. For $\lambda \le 128$ the standard deviations are negligible, then grow quickly (see Figure~\ref{fig:conv:dist-std}). Unreliable values are indicated with gray font. We add $\Bin_{>0}(3,1/2)$ for comparison.}
%. For $\lambda \le 128$, the standard deviation of all $P^*(k \mid n=3, \lambda)$ are negligible. This changes for $\lambda\ge 258$, where the standard deviation becomes important (see main text); these values are therefore printed in gray font and should not be used in an algorithm. }
    \centering\small\setlength{\tabcolsep}{2pt}
    \newcommand{\bad}[1]{\textcolor{gray}{#1}}
    \begin{tabular}{c|cccccccc|c}
        \toprule
        & \multicolumn{8}{c|}{$\lambda$} & \\
        $k$ & $2^3$ & $2^4$ & $2^5$ & $2^6$ & $2^7$ & $2^8$ & $2^9$ & $2^{10}$ & $\Bin_{>0}(3,\frac{1}{2})$ \\
        \midrule
        1 & 0.558 & 0.448 & 0.42937 & 0.42857 & 0.42857 & \bad{0.413} & \bad{0.371} & \bad{0.364} & 0.42857 \\
        2 & 0.331 & 0.414 & 0.42797 & 0.42857 & 0.42857 & \bad{0.408} & \bad{0.364} & \bad{0.342} & 0.42857 \\
        3 & 0.110 & 0.138 & 0.14266 & 0.14286 & 0.14286 & \bad{0.179} & \bad{0.265} & \bad{0.294} & 0.14286 \\
        \bottomrule
    \end{tabular}
    \label{tab:n3}
\end{table}

\begin{figure}[!t]
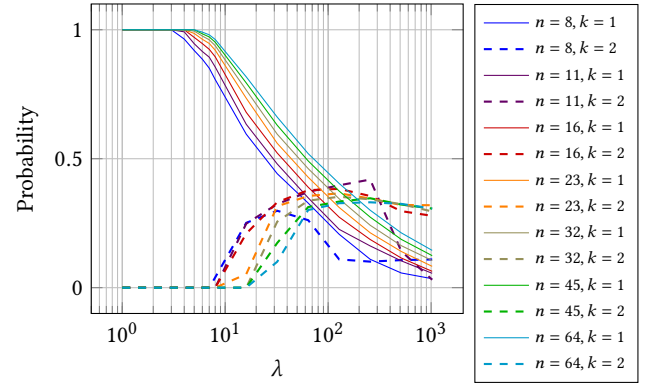

\OneAndTwoBitFlips
\caption{$P^*(1 \mid \lambda, n)$ and $P^*(2 \mid \lambda, n)$ in dependence of $\lambda$, for selected problem dimensions $n$.}
\label{fig:k12}
\end{figure}

\textbf{$\bm{P^*(1 \mid \lambda, n)}$ decreases monotonically with increasing $\bm{\lambda}$.} For fixed dimension $n$, the importance of 1-bit flips significantly decreases as $\lambda$ grows. We have seen this for $n=16$ in Fig.~\ref{fig:16-lines}. This trend generalizes to other problem dimensions, which can be seen in Fig.~\ref{fig:k12}, where we plot the probabilities $P^*(1 \mid \lambda, n)$ and $P^*(2 \mid \lambda, n)$ for $n \in \{8, 11, 16, 23, 32, 45, 64\}$.

% MB: I think we should omit this one.
% The main statement follows from two previous ones, e.g. for the binomial distribution P(2) > P(1) holds.
% Whether P(2) can or cannot increase back again after decreasing, is a difficult question. I would postpone that discussion until we know more...

%\paragraph{For all $n$, there are values of $\lambda$ for which $P^*(2 \mid \lambda, n)>P^*(1 \mid \lambda, n)$:}
%Another observation that we can make from Fig.~\ref{fig:k12} is that for all dimensions $n$, there are cases for which $P^*(1 \mid \lambda, n)<P^*(2 \mid \lambda, n)$. The values of $\lambda$ at which $P^*(2 \mid \lambda, n)$ exceeds $P^*(1 \mid \lambda, n)$ for the first time are non-monotonic in $n$. For example, it is $\lambda=512$ for $n=8$, $\lambda=64$ for $n=10$, $\lambda=128$ for $n \in \{16, 20, 32 \}$, and $\lambda=256$ for $n=64$. The case $n=10$ also shows that the relationship between the two probabilities is not monotonic: for $\lambda=512$ and $\lambda=1024$, the optimal probability to flip one bit is strictly larger again as that of flipping two bits. 

\textbf{$\bm{P^*(2 \mid \lambda, n)}$ is non-monotonic in $\bm{\lambda}$.} We clearly see from Fig.~\ref{fig:k12} that, for fixed dimension $n$, the optimal probability to flip two bits is non-monotonic in $\lambda$. The $\lambda$ at which it becomes non-zero appears to be monotonic in $n$, however, the set of $\lambda$ we used is not enough to determine the exact threshold. It is $\lambda=8$ for $n = 8$, $8 < \lambda \le 16$ for $n \in \{11,16,23\}$, and it is $16 < \lambda \le 32$ for $n \in \{32, 45, 64, 91, 100\}$. Surprisingly enough, this threshold is always larger than the value at which $P^*(1 \mid \lambda,n)$ becomes less than one. We do not see any pattern in the points at which $\Pr[k=2]$ starts to decrease again.%; e.g., it is somewhere in the range $[128, 256]$ for $n=8$, in the range $[64,128]$ for $n=20$, and in the range $[256, 512]$ for $n=64$. 

\textbf{Flipping all bits can be optimal.} Intuitively, flipping more than $n/2$ bits can be optimal in an elitist algorithm only when $\OM(x)<n/2$. It may therefore be surprising that even for mutation strength $\ell=n$ (i.e., flipping all bits) the optimal static probability can be non-zero already for comparatively small $\lambda$. In Tab.~\ref{tab:knnonzero} we show for which combinations of $n$ and $\lambda$ the optimal probability of flipping all bits is non-zero. Note that for some $n$ we have not seen any $\lambda$ where this probability is nonzero, and this may be related to the parity of $n$: for $n \in \{23, 45, 91\}$ the small $\lambda$ feature rather a nonzero $P^*(n-1 \mid n, \lambda)$ instead. So far we do not have any explanation for the observed patterns.

\begin{table}[!t]
\caption{Combinations of $n$ and $\lambda$ for which $P^*(n \mid n, \lambda)>0$ , i.e., the optimal probability of flipping all bits is non-zero.}\label{tab:knnonzero}\footnotesize
\begin{tabular}{l|lllllllllllll}
\toprule
   & \multicolumn{13}{c}{$\lambda$}                              \\
$n$  & 3 & 4 & 5 & 6 & 7 & 8 & 16 & 32 & 64 & 128 & 256 & 512 & 1024 \\
\midrule
8    &   & + & + & + & + & + &    &    & +  &  +  &  +  &  +  &   +  \\
11   &   &   &   &   &   & + &    &    &    &  +  &  +  &  +  &   +  \\
16   &   &   & + & + & + & + &    &    &    &     &  +  &     &      \\
23   &   &   &   &   &   &   &    &    &    &     &     &     &      \\
32   &   &   & + & + & + & + &    &    &    &     &     &     &   +  \\
45   &   &   &   &   &   &   &    &    &    &     &     &     &      \\
64   &   &   &   & + & + & + &    &    &    &     &     &     &      \\
91   &   &   &   &   &   &   &  + &    &    &     &     &     &      \\
100  &   &   &   & + & + &   &  + &    &    &     &     &     &      \\
\bottomrule
\end{tabular}
\end{table}

\begin{figure}[!t]
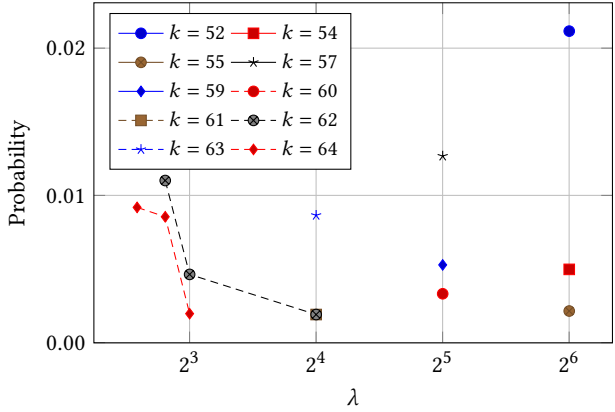

\LargeFlipMusings
\caption{$P^*(k \mid n=64, \lambda)$, in dependence of $\lambda$, and for all values $k \in [50,64]$ for which there is at least one $\lambda$ such that $P^*(k \mid n, \lambda)>0$. Note that, for all these $k$, $P^*(k \mid n, \lambda)=0$ for $\lambda \in \{128, 256, 512, 1024\}$.}
\label{fig:64-tail}
\end{figure}

We take a set $\Lambda = \{6, 7, 8, 16, 32, 64\}$ and show an example in Fig.~\ref{fig:64-tail}, where we display $P^*(k \mid n=64, \lambda \in \Lambda)$ for all $k \in [50,64]$ for which there exists at least one $\lambda\in\Lambda$ with $P^*(k \mid n=64, \lambda) >0$. Interestingly,  $P^*(k \mid n=64, \lambda) =0$ for all tested $\lambda>64$. We also see that only nine different $k$ appear out of $[50,64]$, of which at most three have a non-zero optimal probability for any of the tested $\lambda$.  
%\carola{what else to say here? It's counter-intuitive indeed, but I do not yet know how to best discuss this...}

\textbf{The number of mutation strengths $\bm{k}$ with $\bm{P^*(k \mid \lambda, n)>0}$ increases with~$\bm{\lambda}$, but not monotonically.} We summarize in Tab.~\ref{tab:pos-k} the number of different mutation strengths $k$ for which $P^*(k \mid \lambda, n)>0$. While there is a general trend for an increasing number of such $k$ with increasing $\lambda$, these trends are non-monotonic. From the previous insights, however, it is clear that for every $n$ there exists a threshold $\lambda(n)$ such that for all $\lambda>\lambda(n)$ the number of $k$ with $P^*(k \mid \lambda, n)>0$ is equal to $n$.   

\begin{table}[!t]
\caption{Number of different mutation strengths $k$ for which $P^*(k \mid \lambda, n)>0$, in dependence of $n$ and $\lambda$.}
\label{tab:pos-k}
\footnotesize
\newcommand{\W}[2]{\cellcolor{green!#2!white} #1}
\begin{tabular}{l|lllllllllllll}
\toprule
   & \multicolumn{13}{c}{$\lambda$}                              \\
$n$  &3     &4     &5     &6     &7     &8     &16    &32    &64    &128   &256   &512   &1024 \\
\midrule
8 &\W{1}{7} & \W{2}{14} & \W{2}{14} & \W{3}{21} & \W{3}{21} & \W{5}{35} & \W{4}{28} & \W{5}{35} & \W{7}{49} & \W{8}{56} & \W{8}{56} & \W{8}{56} & \W{8}{56}\\
11 &\W{1}{7} & \W{2}{14} & \W{2}{14} & \W{2}{14} & \W{2}{14} & \W{4}{28} & \W{5}{35} & \W{5}{35} & \W{4}{28} & \W{7}{49} & \W{9}{63} & \W{11}{77} & \W{11}{77}\\
16 &\W{1}{7} & \W{1}{7} & \W{2}{14} & \W{3}{21} & \W{3}{21} & \W{4}{28} & \W{6}{42} & \W{4}{28} & \W{5}{35} & \W{5}{35} & \W{9}{63} & \W{9}{63} & \W{10}{70}\\
23 &\W{1}{7} & \W{1}{7} & \W{2}{14} & \W{2}{14} & \W{2}{14} & \W{4}{28} & \W{5}{35} & \W{6}{42} & \W{6}{42} & \W{6}{42} & \W{7}{49} & \W{9}{63} & \W{11}{77}\\
32 &\W{1}{7} & \W{1}{7} & \W{2}{14} & \W{2}{14} & \W{3}{21} & \W{4}{28} & \W{5}{35} & \W{7}{49} & \W{6}{42} & \W{5}{35} & \W{7}{49} & \W{7}{49} & \W{9}{63}\\
45 &\W{1}{7} & \W{1}{7} & \W{1}{7} & \W{2}{14} & \W{2}{14} & \W{3}{21} & \W{5}{35} & \W{7}{49} & \W{6}{42} & \W{7}{49} & \W{8}{56} & \W{8}{56} & \W{8}{56}\\
64 &\W{1}{7} & \W{1}{7} & \W{1}{7} & \W{2}{14} & \W{3}{21} & \W{4}{28} & \W{4}{28} & \W{7}{49} & \W{6}{42} & \W{8}{56} & \W{8}{56} & \W{7}{49} & \W{8}{56}\\
91 &\W{1}{7} & \W{1}{7} & \W{1}{7} & \W{2}{14} & \W{2}{14} & \W{3}{21} & \W{6}{42} & \W{6}{42} & \W{6}{42} & \W{7}{49} & \W{7}{49} & \W{9}{63} & \W{9}{63}\\
100 &\W{1}{7} & \W{1}{7} & \W{1}{7} & \W{2}{14} & \W{3}{21} & \W{2}{14} & \W{6}{42} & \W{6}{42} & \W{6}{42} & \W{7}{49} & \W{8}{56} & \W{9}{63} & \W{9}{63}\\
\bottomrule
\end{tabular}
\end{table}

\textbf{A closer look for small $\bm{\lambda}$.} It is well known that small values of $\lambda$ are preferable for optimization of simple functions such as \onemax~\cite{JansenJW05}. %and \lo~\cite{DoerrYR0B18}. 
We therefore take a more detailed look at small values in Tab.~\ref{tab:n10-Lsmall-table}, where we list for $n=11$, $\lambda \in [4..8] \cup \{ 16, 32, 64\}$, and for all possible mutation strengths $k \in [11]$ the optimal probability $P^*(k \mid n, \lambda)$. We recall from above that for 
$\lambda\le 3$ the one-point distribution assigning all weight to $k=1$ is optimal. For $\lambda=4$ and $\lambda=7$, the optimal probability $P^*(1 \mid n=11, \lambda)$ is slightly less than 1, and the remaining probability mass is assigned to $k=10$, i.e., to the operator flipping all bits but one. For $\lambda = 8$ two more values, $k=4$ and $k=11$, are also chosen with positive probability. As already discussed in the context of Tab.~\ref{tab:pos-k} and Fig.~\ref{fig:n11-large}, the number of mutation strengths $k$ with $P^*(k \mid n=11, \lambda)>0$ increases for increasing $\lambda$, and the distribution converges to the conditional binomial distribution $\Bin_{>0}(11,1/2)$, which we include in the table for reference. 

\begin{table}[!t]
\caption{Values of $P^*(k \mid n=11,\lambda)$ for selected $\lambda$.}
\label{tab:n10-Lsmall-table}
\newcommand{\W}[2]{\cellcolor{green!#2!red!50!white} #1}
\footnotesize\setlength{\tabcolsep}{2pt}
\begin{tabular}{r|rrrrrrrr|r}
\toprule
& \multicolumn{8}{c|}{$\lambda$} & \\
$k$ & 4 & 5 & 6 & 7 & 8 & 16 & 32 & 64 & $\Bin_{>0}(11, \frac{1}{2})$\\
\midrule
1 & \W{0.9919}{99} & \W{0.9459}{94} & \W{0.9156}{91} & \W{0.8937}{89} & \W{0.8555}{85} & \W{0.6337}{63} & \W{0.4797}{47} & \W{0.3483}{34} & \W{0.0054}{0} \\
2 & & & & & & \W{0.2483}{24} & \W{0.3241}{32} & \W{0.3714}{37} & \W{0.0269}{2} \\
3 & & & & & & & & & \W{0.0806}{8} \\
4 & & & & & \W{0.0530}{5} & & & & \W{0.1612}{16} \\
5 & & & & & & & & & \W{0.2257}{22} \\
6 & & & & & & & & & \W{0.2257}{22} \\
7 & & & & & & & \W{0.1068}{10} & \W{0.1865}{18} & \W{0.1612}{16} \\
8 & & & & & & \W{0.0162}{1} & \W{0.0677}{6} & \W{0.0938}{9} & \W{0.0806}{8} \\
9 & & & & & & \W{0.0748}{7} & \W{0.0217}{2} & & \W{0.0269}{2} \\
10 & \W{0.0081}{0} & \W{0.0541}{5} & \W{0.0844}{8} & \W{0.1063}{10} & \W{0.0875}{8} & \W{0.0270}{2} & & & \W{0.0054}{0} \\
11 & & & & & \W{0.0040}{0} & & & & \W{0.0005}{0} \\
\bottomrule
\end{tabular}
\end{table}

\section{Runtime Comparison}
\label{sec:comparison}

After having focused on the distributions in the previous section, we now study the runtime of the optimal $(1+\lambda)$~UUSD-EA in comparison to other common $(1+\lambda)$~UUSD-EAs. We include in our comparison the \opl variants with $\SBM$, $\SBM_{>0}$, and $\SBM_{0 \rightarrow 1}$ standard bit mutation operators, the  $(1+\lambda)$~RLS, and the $(1+\lambda)$~fastGA with different $\beta\in \{1.3, 1.5, 1.7\}$. We also considered the variant of the latter which directly samples mutation strengths $\ell$ proportional to $\ell^{-\beta}$ for the same parameter values.

 It is not difficult to see that, for any fixed $n$, the expected runtime of the optimal $(1+\lambda)$~UUSD-EAs converge to $1-1/2^n$ as $\lambda \to \infty$. This is also the case for all $(1+\lambda)$~UUSD-EA variants that assign a positive probability to each positive mutation strength $k \in [n]$, and this for all problems $f:\{0,1\}^n \to \R$. 
 Since SBM and fast mutation satisfy these requirements, the expected runtime of all \opl variants as well as that of the fastGA variants also converges to $1-1/2^n$, but at a possibly much different speed. The expected runtime of the generalized $(1+\lambda)$~RLS, in contrast, converges to $n/2$ on \om, since, hand-waivingly, this is the expected distance to the optimum after initialization and the probability to make a progress of 1 in each iteration converged to 1 as $\lambda \to \R$. 
 
\begin{table*}[!h]
\caption{Expected runtimes (in generations) of the optimal $(1+\lambda)$~UUSD-EA(s) on \om for different combinations of $n$ (rows) and $\lambda$ (columns), rounded to two digits. 
For every fixed $n$, the optimal expected runtime converges to $1-1/2^n$ as $\lambda \to \infty$.
}
\label{tab:ET}
\newcommand{\phz}{\phantom{0}}
\newcommand{\phy}{\phantom{0!}}
\newcommand{\pht}{\phantom{!}\!}
\centering
\begin{tabular}{r|rrrrrrrrrrrrrrr}
\toprule
    & \multicolumn{15}{c}{$\lambda$} \\
n   & 1\phy & 2\phy & 3\phy & 4\phy & 5\phy & 6\phy & 7\phy & 8\phy & 16\phz & 32\phz & 64\phz & 128\pht & 256\pht & 512\pht & 1024 \\
\midrule
3   & 3.50   & 2.26   & 1.87   & 1.64   & 1.48   & 1.37   & 1.28  & 1.21  & 0.96  & 0.88  & 0.88  & 0.88  & 0.87  & 0.87  & 0.87  \\
5   & 7.97   & 4.78   & 3.78   & 3.26   & 2.92   & 2.67   & 2.49  & 2.35  & 1.78  & 1.39  & 1.10  & 0.98  & 0.97  & 0.97  & 0.97  \\
8   & 16.20  & 9.32   & 7.12   & 6.04   & 5.36   & 4.89   & 4.54  & 4.27  & 3.15  & 2.43  & 1.98  & 1.67  & 1.39  & 1.14  & 1.01  \\
11  & 25.59  & 14.45  & 10.84  & 9.11   & 8.04   & 7.30   & 6.76  & 6.34  & 4.63  & 3.53  & 2.82  & 2.33  & 2.01  & 1.83  & 1.62  \\
16  & 43.00  & 23.87  & 17.64  & 14.62  & 12.83  & 11.60  & 10.71 & 10.02 & 7.26  & 5.46  & 4.31  & 3.54  & 3.01  & 2.61  & 2.26  \\
23  & 69.95  & 38.35  & 28.02  & 22.99  & 20.04  & 18.04  & 16.60 & 15.50 & 11.14 & 8.32  & 6.51  & 5.30  & 4.46  & 3.83  & 3.34  \\
32  & 107.69 & 58.52  & 42.40  & 34.52  & 29.91  & 26.84  & 24.62 & 22.94 & 16.34 & 12.14 & 9.42  & 7.62  & 6.38  & 5.47  & 4.79  \\
45  & 166.58 & 89.83  & 64.63  & 52.27  & 45.03  & 40.25  & 36.82 & 34.23 & 24.17 & 17.84 & 13.74 & 11.05 & 9.21  & 7.87  & 6.86  \\
64  & 259.25 & 138.90 & 99.31  & 79.86  & 68.44  & 60.95  & 55.59 & 51.56 & 36.09 & 26.45 & 20.23 & 16.17 & 13.39 & 11.41 & 9.92  \\
91  & 400.44 & 213.38 & 151.76 & 121.44 & 103.60 & 91.93  & 83.62 & 77.39 & 53.70 & 39.07 & 29.70 & 23.59 & 19.45 & 16.50 & 14.31 \\
100 & 449.42 & 239.17 & 169.88 & 135.79 & 115.70 & 102.57 & 93.24 & 86.24 & 59.70 & 43.36 & 32.90 & 26.09 & 21.48 & 18.22 & 15.79 \\
\bottomrule
\end{tabular}
\end{table*}

In Tab.~\ref{tab:ET} we present the expected runtimes of the optimal $(1+\lambda)$~UUSD-EA(s) on \om for all different combinations of $n$ and $\lambda$ we have.
Note that these numbers are the lower bounds for all $(1+\lambda)$~UUSD-EAs, including the algorithms mentioned above. Note also that algorithms obtaining a better expected runtime require adaptive parameter choices. 
 
\begin{figure}[!t]
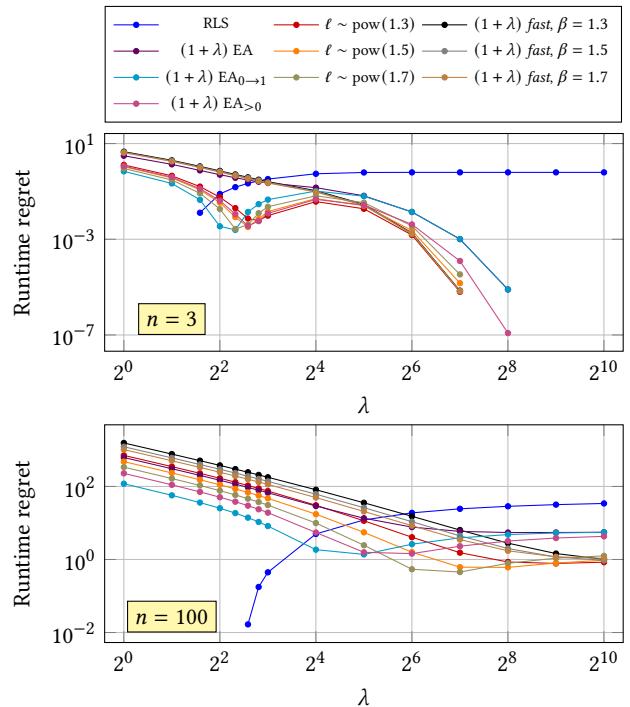

\RuntimeComparisonsThree\par
\RuntimeComparisonsHundred
\caption{Regrets of expected runtime of different $(1+\lambda)$~UUSD-EAs on \om, in dependency of $\lambda$, compared to the runtime of the optimal $(1+\lambda)$~UUSD-EA computed in Sec.~\ref{sec:distributions}, for $n=3$ (top) and $n=100$ (bottom). The algorithms that sample the mutation strength $\ell$ directly from the power-law distribution with parameter $\beta$ are denoted as $\ell\sim\text{pow}(\beta)$. The notation for other algorithms is standard.}
\label{fig:RT-n3}\label{fig:RT-n100}
\end{figure}

In Fig.~\ref{fig:RT-n3} we illustrate the regret in the expected runtime of common $(1+\lambda)$~UUSD-EAs compared to the  optimal one, for $n\in\{3,100\}$, respectively. Corresponding to our discussion above, we observe that for $n=3$ all algorithms, with the exception of RLS, converge to the same optimal expected runtime of $0.875=1-1/8$ when $\lambda \to \infty$, whereas the generalized $(1+\lambda)$~RLS converges to $1.5$, which corresponds to a factor $12/7 \approx 1.71...$ relative disadvantage.

Note that the regrets are not monotone for those algorithms which never flip zero bits, except for the $(1+\lambda$ RLS.
As $\lambda$ grows from the small values, their regret decreases, most probably as flipping more than one bit becomes a better choice. However, with further increase of $\lambda$, the fact that these algorithms flip many bits with a small probability turns to a disadvantage. Note how the heavy-tailed algorithm that samples the mutation strength directly from a power-law distribution with the smallest tested $\beta=1.3$ becomes a clear winner at $\lambda \ge 7$. The similar behavior is seen also for $n=100$ with the exception that the values of $\lambda$ are not large enough to observer the complete convergence picture.

% MB: It is not really different, and given that we don't see the whole picture I don't think the numbers mean something at this point
%The different convergence behavior of the expected runtime of the different algorithms against the optimal one can be observed in Fig.~\ref{fig:RT-n100}. For $\lambda=1\,024$, the relative disadvantage of the fastGA variants against the optimal  $(1+\lambda)$~UUSD-EA(s) is between 5\% (for $\beta=1.3$) and 8\% (for $\beta=1.7$), whereas the disadvantage of the \opl variants is between $27\%$ (for the \opl using $\SBM_{>0}$) and $35\%$ (for the \opl using classic SBM and the one using $\SBM_{\shift}$).

% MB: I think I partially addressed it in the preceding paragraph.
%\carola{We should say a few words about medium $\lambda$. For small $\lambda$, we know that RLS is optimal. It then starts to get worse when $\lambda$ is large enough so that the probability of obtaining fitness increases of two or more in the later parts of the optimization process becomes large enough to outperform the conservative 1-bit flips of RLS.  }

%%%%%%%%%%%%%%%%%%%%%%%%%%%%%%%%%%%%%%%%%%%%%%%%%%%%%%%%%%%
%%                      Conclusion                       %%
%%%%%%%%%%%%%%%%%%%%%%%%%%%%%%%%%%%%%%%%%%%%%%%%%%%%%%%%%%%

\section{Conclusion}
\label{sec:conclusion}

We have analyzed in this paper the dependence of $P^*(k \mid n, \lambda)$, the optimal probability of flipping $k$ bits in the $(1+\lambda)$~EA-UUSD, in dependence of $n$ and $\lambda$. Among other insights, we have shown that the $(1+\lambda)$~RLS is optimal when $\lambda$ is small, and that the value for which it ceases to be optimal increases with increasing $n$. We have also seen that, for fixed $n$, the distribution  $P^*(k \mid n, \lambda)$ converges to the conditional binomial distribution $\Bin_{>0}(n,1/2)$ when $\lambda \to \infty$.

For future work, we consider the following particularly exciting.

\textbf{1) Formalizing the observations into rigorous results:} 
We believe that some of the observations made in this work could be formalized with reasonable effort.

\textbf{2) Analyzing benefits of generalized mutation for more complex problems:} For practitioners, our work is perhaps most interesting in that it invites to consider mutation operators through the lens of probability distributions over the set of possible radii. This idea should show its full potential on problems that are more complex than \om. 
The fastGA proposed in~\cite{fastGA} and fast AIS from~\cite{fast-ais} are compelling examples that show that such generalization can indeed be very beneficial~\cite{MironovichB17,NGopt}.

\textbf{3) Transferring the generalizations to variation operators of higher arity:} The quest for analyzing more general variation operators is not restricted to mutation alone, but also generalizes to variation operators of higher arity, called ``crossover'' or ``recombination'' operators in evolutionary computation. In a simplest extension, one could study effects of changing the binomial distribution associated with the number of bits that is taken from each parent in uniform crossover. We did not yet investigate this idea further, but we hope that a de-coupling of mean and variance, similarly as proposed for variation in~\cite{YeDB19}, may be beneficial. % MB: but I personally think crossovers are much harder than anything with these two words.

\textbf{4) Interplay of generalized mutation with other variation operators:} 
The benefits of generalized mutation operators are very likely not restricted to mutation-only algorithms, but could also improve algorithms that use variation operators of different arities. First examples demonstrating clear advantages of heavy-tailed mutation in the $(1+(\lambda,\lambda))$~GA~\cite{DoerrDE15} were recently shown in~\cite{AntipovD20jump,AntipovBD20fastLLGA}.

\textbf{5) Extensions to the dynamic case:} We studied in this work the case of \emph{static} distributions $P^*(k \mid n, \lambda)$, whereas it is well known that a dynamic choice of the mutation rates, or algorithms' parameters in general, can lead to significant performance gains~\cite{KarafotiasHE15,DoerrD18chapter}. Combining the analyses made in~\cite{BuzdalovD20} for $\flipOp{k}$ with the optimal dynamic $k$ and the optimal dynamic SBM operators with the approach taken in this work (the  generalization to arbitrary distributions) would provide an exact quantification of the disadvantage of static against dynamic mutation operator choices.

% MB: I have nothing in particular to say here, as it requires more insights from constrained optimization, which I am not proficient in, and I got somewhat tired when meeting the page limit.

%\textbf{6) Extension to larger dimensions:} Last, but not least, the results described in this work are comparatively limited in the dimension of the problem. To gather additional intuition about the optimal distributions $P^*(k \mid n, \lambda)$, an extension to larger dimensions would be desirable. \carola{Maxim, could you add a word or two on what's needed for this}

% \carola{Larger alphabets could also be fun, but let's keep this for an extension}
%\carola{We should not forget to add acks for arxiv and final version}

%%%%%%%%%%%%%%%%%%%%%%%%%%%%%%%%%%%%%%%%%%%%%%%%%%%%%%%%%%%
%%                      Bibliography                     %%
%%%%%%%%%%%%%%%%%%%%%%%%%%%%%%%%%%%%%%%%%%%%%%%%%%%%%%%%%%%

\bibliographystyle{ACM-Reference-Format}
\bibliography{paper}

\end{document}